%% file: main.tex
\documentclass[runningheads]{llncs}

\usepackage{eccv}

\usepackage{eccvabbrv}

\usepackage{graphicx}
\usepackage{booktabs}

\usepackage[accsupp]{axessibility}  % Improves PDF readability for those with disabilities.
\usepackage{multirow}
\usepackage{soul}
\usepackage{url}
\usepackage{color}
\usepackage{balance} 
\usepackage{xspace}
\usepackage{tabularx}
\usepackage{colortbl}
\usepackage{enumitem}
\usepackage{wasysym}
\usepackage{pifont}
\usepackage{amsmath}
\usepackage{algorithm}
\usepackage{algpseudocode}
\usepackage{marvosym} 

\definecolor{bblue}{rgb}{0,150,230}
\definecolor{mygray}{gray}{.9}
\definecolor{lightgray}{gray}{.96}
\definecolor{myy}{RGB}{126,95,0}
\definecolor{ggray}{RGB}{127,127,127}
\definecolor{mygreen}{RGB}{93,173,85}
\definecolor{myred}{RGB}{240,16,89}
\definecolor{myblue}{RGB}{0,114,188}
\definecolor{darkgreen}{rgb}{0.0, 0.5, 0.0}
\definecolor{demphcolor}{RGB}{100,100,100}

\usepackage[pagebackref,breaklinks,colorlinks]{hyperref}
\usepackage{orcidlink}

\begin{document}

% ---------------------------------------------------------------
% TODO REVIEW: Replace with your title
\title{AGLLDiff: Guiding Diffusion Models Towards Unsupervised Training-free Real-world \\Low-light Image Enhancement} 

% TODO REVIEW: If the paper title is too long for the running head, you can set
% an abbreviated paper title here. If not, comment out.
\titlerunning{AGLLDiff}

% TODO FINAL: Replace with your author list. 
% Include the authors' OCRID for the camera-ready version, if at all possible.
\author{Yunlong Lin\scalebox{0.75}{*}\inst{1}\orcidlink{0000-0002-3882-158X} \and
Tian Ye\scalebox{0.75}{*}\inst{2}\orcidlink{0000-0002-8255-2997} \and
Sixiang Chen\scalebox{0.75}{*}\inst{2}\orcidlink{0009-0003-6837-886X} \and 
Zhenqi Fu\inst{4}\orcidlink{0000-0003-2950-7190}\and \\
Yingying Wang\inst{1}\orcidlink{0009-0004-3748-1191}\and
Wenhao Chai\inst{5}\orcidlink{0000-0003-2611-0008} \and
Zhaohu Xing\inst{2}\orcidlink{0009-0002-2502-3578}  \and
Lei Zhu\inst{2,3}\orcidlink{0000-0003-3871-663X}\and
Xinghao Ding\inst{1}\orcidlink{0000-0003-2288-5287}\textsuperscript{\Letter}}

% TODO FINAL: Replace with an abbreviated list of authors.
\authorrunning{Lin et al.}
% First names are abbreviated in the running head.
% If there are more than two authors, 'et al.' is used.

% TODO FINAL: Replace with your institution list.
\institute{Xiamen University, China \and
The Hong Kong University of Science and Technology (Guangzhou), China \and
The Hong Kong University of Science and Technology,  Hong Kong SAR, China \and
Tsinghua University, China \and
University of Washington \\
\email{dxh@xmu.edu.cn}\\
{\tt\small Project page: \url{https://aglldiff.github.io/}}
% \url{http://www.springer.com/gp/computer-science/lncs} \and
% ABC Institute, Rupert-Karls-University Heidelberg, Heidelberg, Germany\\
}

\maketitle
\begin{abstract}
Existing low-light image enhancement (LIE) methods have achieved noteworthy success in solving synthetic distortions, yet they often fall short in practical applications. The limitations arise from two inherent challenges in real-world LIE: 1) the collection of distorted/clean image pairs is often impractical and sometimes even unavailable, and 2) accurately modeling complex degradations presents a non-trivial problem. To overcome them, we propose the Attribute Guidance Diffusion framework (AGLLDiff), a training-free method for effective real-world LIE. Instead of specifically defining the degradation process, AGLLDiff shifts the paradigm and models the desired attributes, such as image exposure, structure and color of normal-light images. These attributes are readily available and impose no assumptions about the degradation process, which guides the diffusion sampling process to a reliable high-quality solution space. Extensive experiments demonstrate that our approach outperforms the current leading unsupervised LIE methods across benchmarks in terms of distortion-based and perceptual-based metrics, and it performs well even in sophisticated wild degradation.
% Extensive experiments demonstrate that our approach can achieve visual-friendly results and perform robustly even in sophisticated wild degradation. 
\end{abstract}

\input{sec/1_intro}
\input{sec/2_related_work}
\input{sec/3_method}
\input{sec/4_Experiments}
\input{sec/5_Conclusions}

{
% ---- Bibliography ----
%
% BibTeX users should specify bibliography style 'splncs04'.
% References will then be sorted and formatted in the correct style.    
    \bibliographystyle{splncs04}
    \bibliography{references}
}

% \input{sec/X_suppl}
% {
%    \small
%    \bibliographystyle{splncs04}
%    \bibliography{references}
% }

\end{document}

%% file: sec/1_intro.tex
\section{Introduction}
\label{sec:intro}
Real-world low-light image enhancement (LIE) aims to ameliorate the quality and brightness of an image suffering from unknown degradation, such as low contrast, multiple artifacts, poor visibility, sensor noise, etc. Great improvement in enhancement quality has been witnessed over the past few years with the exploitation of generative priors~\cite{GAN1,GAN2,Enlightengan}. For instance, Generative Adversarial Networks (GANs)~\cite{GAN3,GAN4} that are trained on extensive datasets of clean images and learn rich knowledge of real-world scenes have succeeded in LIE through GAN inversion. Compared to GANs, Denoising Diffusion Probabilistic Models (DDPMs)~\cite{ddpm,ddim,DDPM1,DDPM2,DDPM3,cao2023difffashion,cao2023image,jiang2024back} yield more high-fidelity and realistic details, thereby fostering a surge of interest in adapting diffusion models to LIE~\cite{CLE,LLDiffusion,jiang2023low,PyramidDiff,GlobalDiff,ReCo-Diff}.

Recent diffusion-based LIE methods can be roughly classified into two categories: \textit{\textbf{1) The common approaches}}~\cite{CLE,LLDiffusion,ReCo-Diff,Diff-retinex} are dedicated to accurately modeling degradation process via supervised learning, which show proficiency in synthetic degradation scenes but lack robustness to handle challenging unseen degradations. This inadaptability primarily stems from the inconsistency between the synthetic degradation of training data and the actual degradation in the real world. Enriching the synthetic data for model training would improve the models' generalizability, but it is obviously impractical to simulate every possible degradation in the real world. \textit{\textbf{2) The second ones}}~\cite{GDP,liu2024diff} strive to exploit the diffusion priors in the pre-trained diffusion models, which are effective in adapting to multiple degradations. Yet, despite their versatility, these methods are inevitably constrained in terms of generalizability, as they require prior knowledge of the specific degradation process in advance. In practice, degradations in the wild often include a mixture of multiple types, posing a challenge to accurately model them. In summary, two primary challenges are commonly encountered in real-world LIE: \textbf{i)} the collection of distorted/clean image pairs is often impractical and sometimes even unavailable, and \textbf{ii)} accurately modeling complex degradations presents a non-trivial problem.
% \textit{\textbf{}}applying diffusion models to
% 1) the collection of distorted/clean image pairs is often impractical and sometimes even unavailable, and 2) accurately modeling complex degradations presents a non-trivial problem.

\begin{figure}[!t]
\centering
\setlength{\abovecaptionskip}{-0.01cm} %调整caption与图的距离
\setlength{\belowcaptionskip}{-0.4cm}%调整caption与下文的距离
\includegraphics[width=1\linewidth]{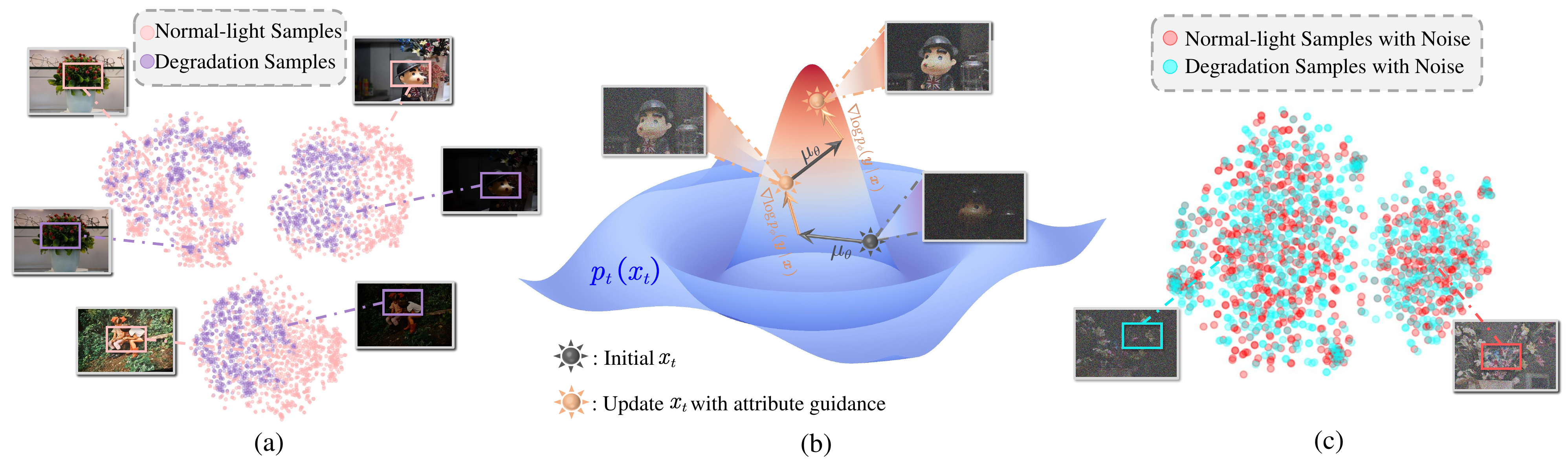}
\caption{\textbf{Motivation of our AGLLDiff.} (a) represents the data distribution of normal-light samples and degraded samples. It is evident that degradation significantly deviates from normal-light samples. (b) conceptually illustrate the geometries of the proposed attribute guidance sampling algorithm. It shows that, given the initial latent, which lies in the low-probability region, attribute guidance guides the latent to move towards its vicinal high-probability region. (c) presents that imposing gaussian noise on the degraded sample and its corresponding reference sample makes the distributions between them less distinguishable.
}\label{intro}
\vspace{-0.2cm}
\end{figure}

To address the aforementioned challenges, we introduce \textit{\textbf{a novel training-free and unsupervised framework, named AGLLDiff}}, for real-world low-light enhancement. In contrast to prior works that predefine the degradation process, \textit{\textbf{our approach models the desired attributes and incorporates this guidance within the diffusion generative process.}} Concrectly, we leverage a well-performing diffusion model (DM)~\cite{diffsuion_beat_gan,stablediff,ControlNet}, which generates images through a stochastic iterative sampling process, and the attributes act as classifiers to constrain the generative process to a reliable high-quality (HQ) solution space. As shown in Fig.~\ref{intro}, noisy images are degradation-irrelevant conditions for the DM generative process. Adding extra gaussian noise makes the degradation less distinguishable compared with its corresponding reference distribution. Since diffusion prior can serve as a natural image regularization, one could simply guide the sampling process with easily accessible attributes such as image exposure, structure and color of normal-light images. \textit{\textbf{By constraining a reliable HQ solution space, the core of our philosophy is to bypass the difficulty of discerning the prior relationship between low-light and normal-light images, thus improving generalizability.}}

Our contributions can be summarized as follows:
\begin{itemize}
\item We introduce a novel paradigm, AGLLDiff, a training-free and unsupervised method that requires no degradation of prior knowledge but yields high fidelity and generality towards real-world low-light image enhancement.
\item We demonstrate that AGLLDiff suffices to guide the pre-trained diffusion models to a reliable high-quality solution space through  easily accessible attributes in the HQ image space.
\item Comprehensive experiments reveal that our framework achieves both robustness and high quality on heavily degraded synthetic and real-world datasets.
\end{itemize}

%% file: sec/2_related_work.tex
\section{Related Work}
% \subsection{Low-light Image Enhancement}
% Enhancing images in low-light conditions has been a longstanding issue and great progress has been made over the decades. 
\textbf{Low-light Image Enhancement.}
To transform low-light images into visually satisfactory ones, numerous efforts have been made over the decades. The conventional approaches are first widely adopted~\cite{LECARM,SDD,cheng2004simple,huang2012efficient,abdullah2007dynamic,rahman2016adaptive}. For example, Wang et al.~\cite{wang2009real} improved the visibility and contrast by applying gamma correction and enhancing the dynamic contrast ratio. Guo et al.~\cite{Guo1} suggested refining the initially estimated illumination map by incorporating a structural prior. Arici et al.~\cite{arici2009histogram} introduced penalty terms to avoid the unnatural look and visual artifacts of the enhanced image. Lee et al.~\cite{DICM} applied the layered difference representation of 2D histograms to amplify the gray-level differences between adjacent pixels. The enhancement performance of current conventional methods relies on tedious hand-crafted priors and is only applicable to specific scenarios.

Recently, the paradigm of low-light image enhancement has gradually shifted to data-driven approaches based on deep learning~\cite{liu2023nighthazeformer,zou2024vqcnir,ZeroDCE,Zerodcepp,pairLIE,SNR,CUE,UHDFourICLR2023,Neco}. For instance, Chen et al.~\cite{RetinexNet} combined Retinex theory with a CNN network to estimate and adjust the illumination map. Ma et al.~\cite{SCI} established a cascaded illumination estimation process to achieve fast and robust LIE in complex scenarios. Lore et al.~\cite{Lore} developed a stacked sparse denoising autoencoder framework aimed at improving the quality of low-light images. Lv et al.~\cite{MBLLEN} presented a multi-branch network that extracts rich features from different levels to enhance low-light images via multiple sub-networks. Xu et al.~\cite{SNR} proposed a signal-to-noise (SNR)-aware network that integrates a convolutional short-range branch with a transformer-based long-range branch. Cai et al.~\cite{Retinexformer} designed a novel one-stage Retinex-based framework for LIE. Additionally, in contexts where training images are scarce, the utility of unsupervised~\cite{pairLIE,Neco} and zero-shot learning approaches~\cite{ZeroDCE,clip-lie,GDP} becomes increasingly pronounced.
% Fu et al.~\cite{Fu} used a weighted variational model to preserve the reflectance with more details. 
% Guo et al.~\cite{Guo1} calculated a well-constructed illumination map for targeted enhancement. Hao et al.~\cite{Hao} proposed a novel Retinex-based LIE method performed in a semi-decoupled way.       
% Lv et al.~\cite{MBLLEN} present a multi branch network, which extracts rich features from different levels, to enhance low-light images via multiple subnets. Wang et al.~\cite{wang2019underexposed} introduce intermediate illumination rather than directly learn an image-to-image mapping. 
% Lv et al.~\cite{MBLLEN} present a multi branch network, which extracts rich features from different levels, to enhance low-light images via multiple subnets. Cai et al.~\cite{Retinexformer} designed a sophisticated transformer-based algorithm for low-light image enhancement and achieved a fresh SOTA performance.
% \subsection{Diffusion-Based Image Restoration and Low-Light Image Enhancement} 
% Diffusion Models (DMs)~\cite{ddim,ddpm,DDRM,DDPM1,DDPM3,DDPM2}, have achieved outstanding results in image generation tasks. DMs adopt parameterized Markov chain to optimize the lower variational bound on the likelihood function, which can make them generate more accurate target distribution than other generative models, i.e., GAN. Recently,

\textbf{Diffusion-Based Image Restoration and Low-Light Image Enhancement.}
Diffusion Models has become increasingly influential in the field of image restoration (IR) tasks~\cite{DDRM,GDP,DDNM,ye2024learning,ye2023adverse,ye2023sequential,chen2023sparse}, such as super resolution~\cite{saharia2022image,zhu2023denoising}, blind face restoration~\cite{Ilvr,wang2023dr2}, image fusion~\cite{lin2023domain,wang2023learning,wang2024cross,huang2023dp}, dehazing~\cite{ye2021perceiving,chen2023dehrformer}, desnowing~\cite{chen2023cplformer,chen2022msp,chen2022snowformer},
image enhancement~\cite{GlobalDiff,Diff-retinex,PyramidDiff,ReCo-Diff}. These methods could be broadly categorized into supervised and unsupervised paradigms.
Supervised-based IR solutions usually rely on large-scale, pre-collected paired datasets to train their models with great success. 
Hou et al.~\cite{GlobalDiff} devised a diffusion-based framework, incorporating a global structure-aware regularization to maintain the intricate details and textures within images. Yi et al.~\cite{Diff-retinex} integrated the diffusion model alongside the Retinex model to enhance low-light images.
Jiang et al.~\cite{jiang2023low} employed wavelet transformation to decrease the input size and a high-frequency restoration module to maintain the details. Wang et al.~\cite{wang2023lldiffusion} and Yin et al.~\cite{CLE} directly utilized the color map as an extra conditional control to preserve the color information. A major challenge is that they implicitly assume training and testing data should be identically distributed. As a result, these methods often deteriorate seriously in performance when testing cases deviate from the pre-assumed distribution.

Another prevailing research line is unsupervised-based IR approaches. They adopt a zero-shot approach to leverage a pre-trained diffusion model for restoration without the need for task-specific training. As an early attempt, Kawar et al.~\cite{DDRM} hypothesized the linear degradation model and relied on the desirable property of linear formulation to sample from posterior distribution. Wang et al.~\cite{DDNM} introduced the range-null space decomposition to further improve the zero-shot image restoration. Fei et al.~\cite{GDP} applied a simultaneous estimation of degradation model to address blind degradation. Yang et al.~\cite{yang2024pgdiff} introduced a partial guidance mechanism for blind face restoration, wherein intermediate outputs of the diffusion model are constrained by a classifier to perform photo restoration.
Previous diffusion-based image restoration methods explicitly leveraged a degradation model by solving a maximum posterior problem or a posterior sampling problem to generate solutions. However, for many practical image enhancement problems, the underlying degradation model may not be available.
\textit{\textbf{In this work, we propose to model the desired attributes of normal-light images. Such a strategy is independent of the degradation process, circumventing the difficulty of modeling the degradation process.}}

% As a classic prepossessing task, the practicality of performance-oriented methods is limited. 

%% file: sec/3_method.tex
\section{Methodology}
\subsection{Preliminaries of Diffusion Models}
% \textbf{Denoising Diffusion Model.} 
The diffusion models~\cite{ddpm,ddim,DDPM1,guo2024versat2i,chai2023stablevideo,bai2023integrating,cao2023difffashion} belong to a category of generative models that operate by incrementally incorporating Gaussian noise into training data and subsequently acquiring the denoiser to restore the data distribution $p(\boldsymbol{x})$ by reversing the process of noise injection.

\underline{\emph{The forward process}} $q\left( \boldsymbol{x}_t\mid \boldsymbol{x}_{t-1} \right)$ transforms an initial image $\boldsymbol{x}_{0}$ into Gaussian noise $\boldsymbol{x}_{T}\sim \mathcal{N}(0,1)$ over $T$ iterations. The following equation can express the process of each iteration in the diffusion:
\begin{equation}\label{eq1}
q\left(\boldsymbol{x}_t \mid   \boldsymbol{x}_{t-1}\right)=\mathcal{N}\left(  \boldsymbol{x}_t ; \sqrt{1-\beta_t}   \boldsymbol{x}_{t-1}, \beta_t   \boldsymbol{I}\right),
\end{equation}
where $\boldsymbol{x}_t$ denotes the noisy image at time-step $t$, $\beta_t$ is the pre-determined scaling factor, and $\mathcal{N}$ represents the Gaussian distribution. Under the reparameterization trick, $\boldsymbol{x}_t$ can be written as:
\begin{equation}
    \boldsymbol{x}_t=\sqrt{\bar{\alpha}_t} \boldsymbol{x}_0+\sqrt{1-\bar{\alpha}_t} \epsilon,
    \label{eq2}
\end{equation}
where $\bar{\alpha}_t:=\prod_{i=1}^t\left(1-\beta_i\right)$ and $\epsilon \sim \mathcal{N}(0,I)$. Then $\boldsymbol{x}_T\sim \mathcal{N}(0,I)$ if $T$ is big enough, usually $T=1000$.

% \underline{\emph{The forward diffusion process}} 
% The model progressively generates images by reversing the forward process. The generative process is also a Gaussian transition with the learned mean

\underline{\emph{The reverse generative process}} of the inference stage, starting from a Gaussian random noise map $\boldsymbol{x}_T\sim \mathcal{N}(0,I)$ and iteratively performing the denoising step until it attains a high-quality output $\boldsymbol{x}_{0}$:
% DM methods perform the sampling of a Gaussian random noise map $\boldsymbol{x}_{T}$, subsequently applying a gradual denoising procedure to $\boldsymbol{x}_{T}$ until it attains a high-quality output $\boldsymbol{x}_{0}$.
\begin{equation}
p_{\boldsymbol{\theta }}\left( \boldsymbol{x}_{t-1}\mid \boldsymbol{x}_t \right) =\mathcal{N}\left( \boldsymbol{x}_{t-1};\boldsymbol{\mu }_{\boldsymbol{\theta }}\left( \boldsymbol{x}_t,t \right) ,\Sigma _{\theta}\mathrm{I} \right) ,
\end{equation}
where variance $\Sigma _{\theta}\mathrm{I}$ can be either time-dependent constants~\cite{ddpm} or learnable parameters~\cite{nichol2021improved}. 
The mean value $\boldsymbol{\mu }_{\boldsymbol{\theta }}\left( \boldsymbol{x}_t,t \right)$ is generally parameterized by a network $\boldsymbol{\epsilon }_{\boldsymbol{\theta }}\left( \boldsymbol{x}_t,t \right)$:
\begin{equation}
    \boldsymbol{\mu }_{\boldsymbol{\theta }}\left( \boldsymbol{x}_t,t \right) =\frac{1}{\sqrt{\alpha}_t}\left( \boldsymbol{x}_t-\frac{\beta _t}{\sqrt{1-\bar{\alpha}_t}}\boldsymbol{\epsilon }_{\boldsymbol{\theta }}\left( \boldsymbol{x}_t,t \right) \right).
\end{equation}
In practice, one can also directly approximate $\boldsymbol{\hat{x}}_0
$ from $\boldsymbol{\mu }_{\boldsymbol{\theta }}\left( \boldsymbol{x}_t,t \right)$:
\begin{equation}
    \boldsymbol{\hat{x}}_0=\frac{\boldsymbol{x}_t}{\sqrt{\bar{\alpha}_t}}-\frac{\sqrt{1-\bar{\alpha}_t}\boldsymbol{\epsilon }_{\boldsymbol{\theta }}\left( \boldsymbol{x}_t,t \right)}{\sqrt{\bar{\alpha}_t}}.
\end{equation}

\textbf{Classifier Guidance.} The classifier guidance is employed to direct an unconditional diffusion model towards achieving conditional generation. Here, $\boldsymbol{y}$ represents the target and $\boldsymbol{p}_{\boldsymbol{\phi }}(\boldsymbol{y}\mid \boldsymbol{x})$ symbolizes a classifier, the conditional distribution is formulated to resemble a Gaussian distribution akin to its unconditional counterpart, but with the mean shifted by $\Sigma _{\boldsymbol{\theta }}\left( \boldsymbol{x}_t,t \right) \boldsymbol{g}$~\cite{diffsuion_beat_gan}:
\begin{equation}
    \boldsymbol{p}_{\boldsymbol{\theta },\boldsymbol{\phi }}\left( \boldsymbol{x}_{t-1}\mid \boldsymbol{x}_t,\boldsymbol{y} \right) \approx \mathcal{N}\left( \boldsymbol{\mu }_{\boldsymbol{\theta }}\left( \boldsymbol{x}_t,t \right) +\Sigma _{\boldsymbol{\theta }}\left( \boldsymbol{x}_t,t \right) \boldsymbol{g},\Sigma _{\boldsymbol{\theta }}\left( \boldsymbol{x}_t,t \right) \right),
\end{equation}
where $\boldsymbol{g}:=\left. \nabla _{\boldsymbol{x}}\log \boldsymbol{p}_{\boldsymbol{\phi }}(\boldsymbol{y}\mid \boldsymbol{x}) \right|_{\boldsymbol{x}=\boldsymbol{\mu }_{\boldsymbol{\theta }}\left( \boldsymbol{x}_t,t \right)}$. The gradient $\boldsymbol{g}$ serves as a guidance that leads the unconditional sampling distribution towards the condition target $\boldsymbol{y}$. 

\begin{figure}[!t]
\centering
\setlength{\belowcaptionskip}{-0.4cm}%调整caption与下文的距离
\includegraphics[width=1\linewidth]{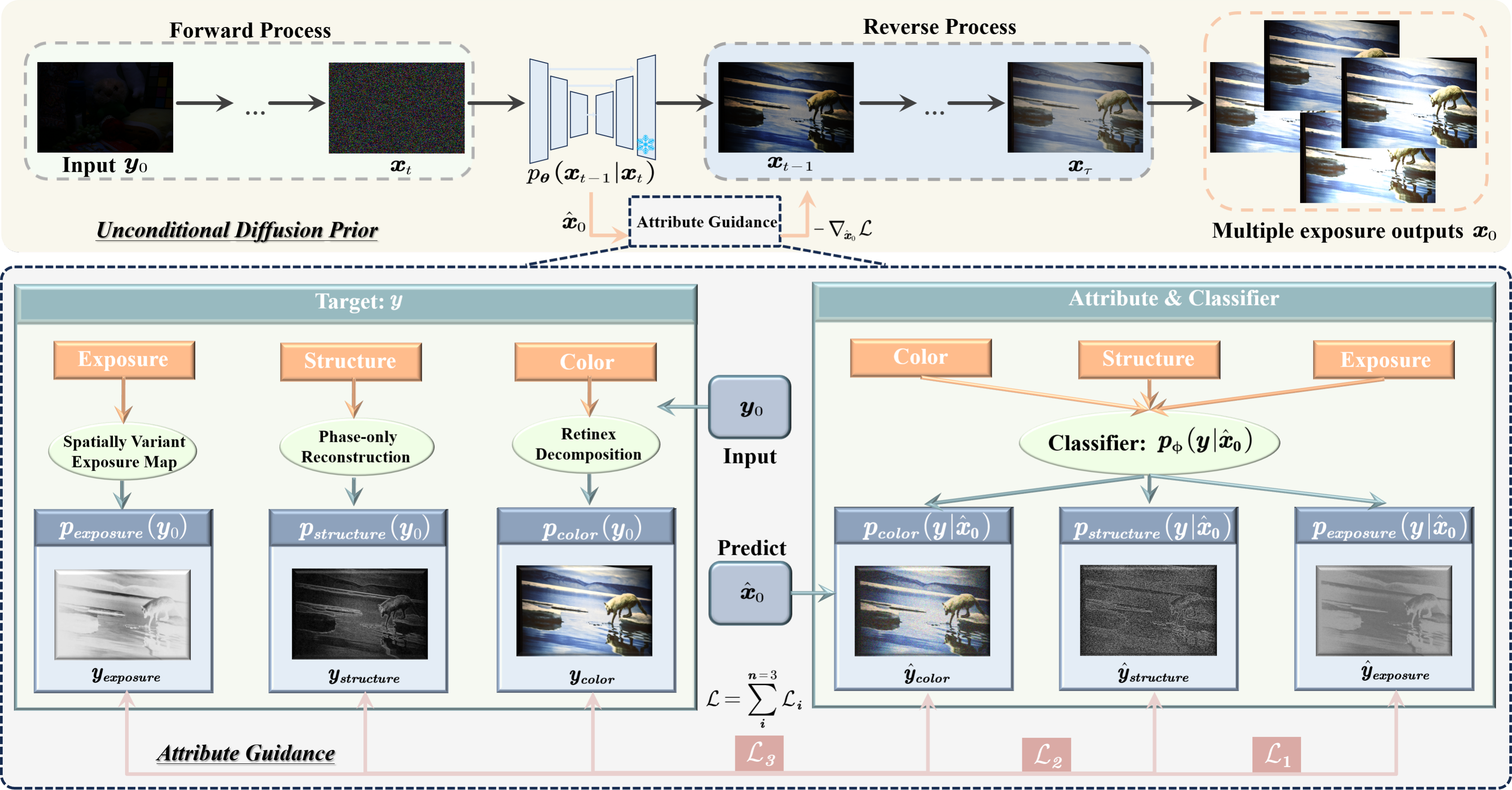}
\caption{The overall framework of our proposed AGLLDiff.
}\label{method}
\end{figure}

\subsection{Overview of the AGLLDiff Framework}
Our core motivation is to model the desired attributes of normal-light images and apply them to guide the diffusion generative process into a reliable high-quality (HQ) space. Such a design is agnostic to the degradation process and bypasses the difficulty of modeling the degradation process, making it more suitable for real-world LIE. The enhanced image should simultaneously satisfy: i) it is faithful to the degraded image, and ii) it conforms to the model distribution of pre-trained diffusion models that incorporate a vast repository of prior knowledge about HQ natural images.
The overview of AGLLDiff is summarized in Fig.~\ref{method} and Algorithm \textcolor{red}{1}. Given a degraded low-light image $\boldsymbol{y}_0$ in the wild domain, the diffusion forward process adds a few steps of slight Gaussian noise to the $\boldsymbol{y}_0$, aiming to narrow the distribution between the degraded image and its potential counterpart, i.e., the HQ image. After obtaining the noisy image $\boldsymbol{x}_t$, we implement the reverse generative process through a pre-trained diffusion denoiser and attribute guidance to generate the enhancement result $\boldsymbol{x}_0$. The inherent attributes of normal-light samples, such as image exposure, structure and color, can be readily derived from their degraded counterparts. Further elaboration on the pivotal components of our method, including attributes, classifiers, and targets, is provided in subsequent sections.

\subsection{Attribute Guidance}\label{Sec3.3}
Our attribute guidance eschews any assumptions about degradation. Instead, with diffusion prior acting as a regularization, we provide guidance only on the desired attributes of HQ images. The key to AGLLDiff is to construct proper guidance on the generative process. 

\begin{figure}[!t]
\centering
\setlength{\abovecaptionskip}{-0.03cm} %调整caption与图的距离
\setlength{\belowcaptionskip}{-0.4cm}%调整caption与下文的距离
\includegraphics[width=1\linewidth]{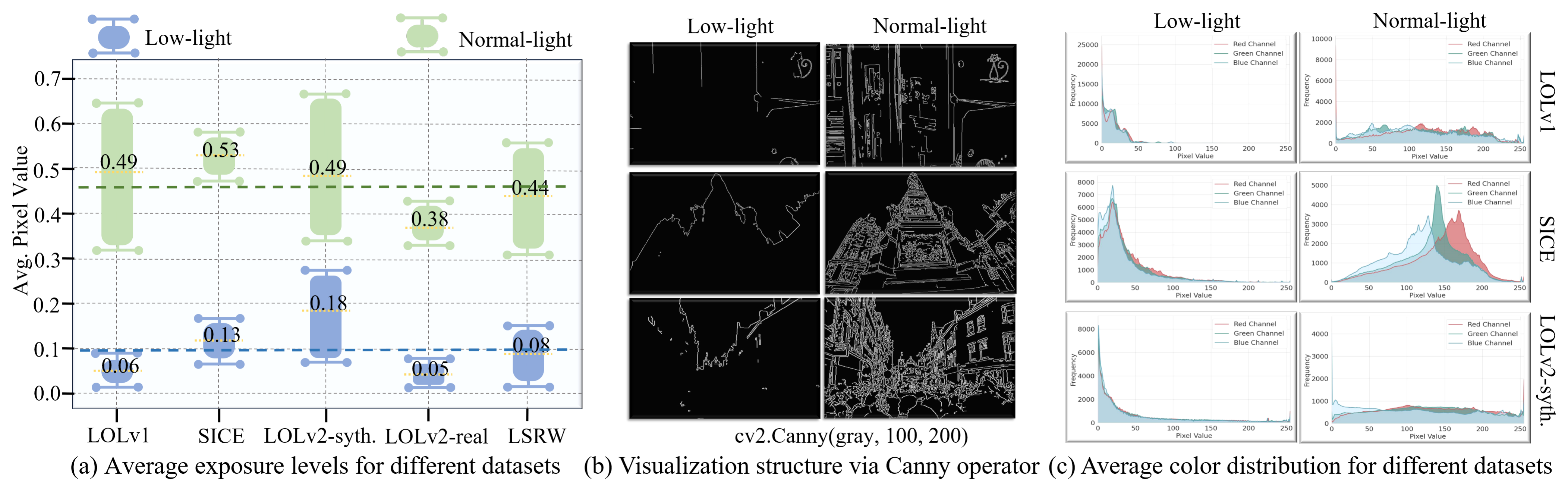}
\caption{\textbf{Statistics and Visualization.} (a) The average exposure values of the low- and normal-light subsets of the five LIE datasets. (b) Visualization of the structure of low- and normal-light images by the Canny operator. (c) Histogram of the color distribution of the low- and normal-light images.
}\label{method2}
\end{figure}

\textbf{Attribute and Classifier.} The initial step of AGLLDiff is to determine the desired attributes that the normal-light output possesses. Each of these attributes corresponds to a specific classifier $\log \boldsymbol{p}_{\boldsymbol{\phi}}(\boldsymbol{y}\mid \boldsymbol{x}_0)$, and the intermediate outputs $\boldsymbol{x}_t$ are updated by back-propagating the gradient computed on the loss between the classifier output and the target $\boldsymbol{y}$. Through this mechanism, enhanced results can be obtained via an iterative refinement. The significance of attributes in achieving the desired final result cannot be overstated. This raises a fundamental question: \textit{\textbf{What attributes are possessed by HQ images? Through comprehensive observation and statistical analysis, we conclude that the following three fundamental attributes are typically found in HQ images: 1) well exposure, 2) clear structure, and 3) vivid colors.}} As presented in Fig.~\ref{method2}, we analyzed the average exposure levels and color distributions of five LIE datasets, visualizing the structure of low- and normal-light images. The observations are as follows: 1) there exists a significant  discrepancy in average exposure values, with low-light images at around 0.1 and normal-light images at approximately 0.46, 2) normal-light images have clearer and richer structures compared to low-light images, and 3) the color distribution of images in normal light is both more colorful and homogeneous than in low-light images. 

\begin{figure}[!t]
\centering
\setlength{\abovecaptionskip}{-0.03cm} %调整caption与图的距离
\setlength{\belowcaptionskip}{-0.4cm}%调整caption与下文的距离
\includegraphics[width=1\linewidth]{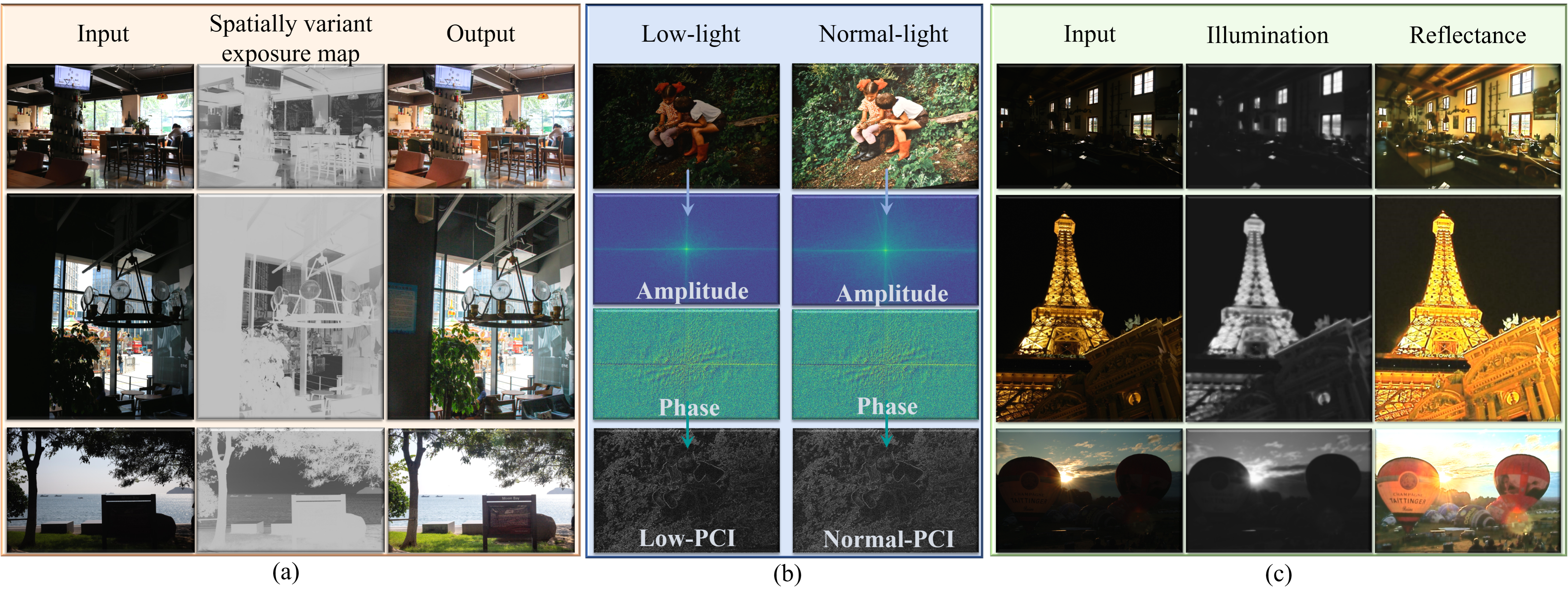}
\caption{(a) Visualization of the spatially variant exposure maps. Based on Eq.~\ref{eq8}, we automatically assign the underexposed regions large exposure values (light gray) and wellexposed/overexposed regions small exposure values (dark gray). (b) Visualization of the phase-only reconstruction image (PCI) in the spatial domain. We applied an inverse discrete Fourier transform to the phase of the low/normal-light image to obtain the phase-only reconstruction image. That means the
amplitude of low/normal-light image is set to 1. (c) Visualization of the Retinex decomposition. We employ a pre-trained decomposition network, RNet, to decompose the input into a reflectance map $R$ and an illumination map $L$.}\label{method3}
\end{figure}

\textbf{Exposure Guidance Formulation.} To guide the output exposure toward that of normal-light images, we utilize the spatially variant exposure map~\cite{li2022cudi} to constrain the exposure of the output $\boldsymbol{\hat{x}}_0$. The loss is formulated as follows:
\begin{equation}
\mathcal{L}_1=\left\| \mathrm{Mean}\left( \boldsymbol{\hat{x}}_0 \right) -\mathrm{Mean}\left( E \right) \right\| _{2}^{2},
\end{equation}
where $E$ denotes the spatially variant exposure map. Specifically, the exposure map is set with non-uniform exposure values in different regions, e.g., the underexposed region is assigned a large exposure value while the overexposed region is allocated a small exposure value. To achieve the spatially variant exposure map, we first obtain the luminance channel $Y$ by color space decomposition (e.g., YCbCr, YUV) and its average value $Y_{avg}$ of a low-light image $\boldsymbol{y}_0$. Then we calculate the spatially variant condition exposure map by:
\begin{equation}
E=A\times \mathrm{Norm}\left( Y-Y_{\mathrm{avg}} \right) +B,
\label{eq8}
\end{equation}
where $B$ is the base exposure value, $A$ is the adjustment amplitude, and $\mathrm{Norm}$ operation normalizes its input to the range of $\left[ -1,1 \right]$. Based on extensive statistics in Fig.~\ref{method2}(a), $B$ and $A$ are empirically set to 0.46 and 0.25, respectively. Fig.~\ref{method3}(a) presents three outcomes of our method utilizing spatially variant exposure maps, where underexposed areas receive higher exposure values and well-exposed or overexposed areas receive lower ones. Such a spatially variant setting enables precise exposure adjustments and allows for generating multiple results at varying exposure levels by adjusting the exposure values.

\textbf{Structure Guidance Formulation.} For constraining the structure of the output faithful to the degraded image, we minimize the phase error between the degraded image $\boldsymbol{y}_0$ and the output $\boldsymbol{\hat{x}}_0$:
\begin{equation}
\mathcal{L}_2=\left\| \mathcal{P}\left( \boldsymbol{\hat{x}}_0 \right) -\mathcal{P}\left( \boldsymbol{y}_0 \right) \right\| _{2}^{2},
\end{equation}
where $\mathcal{P}(\cdot)$ indicates the phase in the Fourier domain. In Fig.~\ref{method3}(b), we perform the inverse discrete Fourier transform to obtain phase-only reconstruction images in the spatial domain. As observed, the phase-only reconstruction versions of low-light and normal-light exhibit structural consistency. This is because most illumination information is expressed as amplitudes, and structural information is revealed in phases~\cite{UHDFourICLR2023}. We find that this simple phase constraint is sufficient to produce reliable results.

\textbf{Color Guidance Formulation.} According to the Retinex theory~\cite{land1971lightness}, a low-light image can be decomposed into illumination $L$ and reflectance $R$. As shown in Fig.~\ref{method3}(c), the reflectance map $R$ represents the physical properties of the objects, which contain abundant color information. Therefore, we could guide the color synthesis process with the reflectance map $R$. Equivalently, the loss is formulated as follows:
\begin{equation}
    \mathcal{L}_3=\left\| \mathcal{F}\left( \boldsymbol{\hat{x}}_0 \right) -\mathcal{F}\left( \boldsymbol{y}_0 \right) \right\| _{2}^{2},
\end{equation}
where $\mathcal{F}\left( \cdot \right)$ denotes the pre-trained Retinex-based decomposition network~\cite{pairLIE}, termed RNet. It takes a low-light image and generates the reflectance map $R$ and illumination map $L$. RNet learns adaptive physical-based constraints from low-light image pairs in a self-supervised manner, significantly reducing the dependence on hand-crafted priors. Such an effective constraint ensures that our method can generalize well to various exposure scenes, which is why we chose it. %More details can be found in \cite{pairLIE}.
% Therefore, we utilize it as 
% Inspired by the training strategy in PairLIE~\cite{pairLIE}, we utilize the reflectance consistency loss, reconstruction loss, smooth illumination loss to pre-train RNet. More details can be seen in the supplementary materials.

Our attribute guidance controls only the attributes of HQ outputs, and therefore composing the classifiers and summing the loss corresponding to each attribute can easily guide the diffusion model to generate HQ results:
\begin{equation}
\mathcal{L}\,\,=\,\,\lambda _1\mathcal{L}_1+\lambda _2\mathcal{L}_2+\lambda _3\mathcal{L}_3,
\end{equation}
where $\lambda _1$, $\lambda _2$ and $\lambda _3$ are constants controlling the relative importance of the different losses, which are empirically set to 1000, 10 and 0.03 in all experiments, respectively.
% . $\lambda _1$, $\lambda _2$ and $\lambda _3$
\begin{algorithm}[!t]
\caption{Sampling with attribute guidance}
\begin{algorithmic}
\State \textbf{Require}: A pre-trained diffusion model $(\boldsymbol{\mu}_{\boldsymbol{\theta}}(\boldsymbol{s}_t, t), \Sigma_{\theta}(\boldsymbol{x}_t, t))$, classifier $\boldsymbol{p}_{\boldsymbol{\theta}}(\boldsymbol{y}|\boldsymbol{x}_0)$, target $\boldsymbol{y}$, gradient scale $\boldsymbol{s}$, the number of gradient steps $\boldsymbol{N}$ and the iteration steps $\boldsymbol{\omega}$ of adding and removing noise.

\State \textbf{Input}: A low-light image $\boldsymbol{y}_0$
\State \textbf{Output}: Output image $\boldsymbol{x}_0$
% \State Input: a low-quality image $y_0$
% \State $x_T \gets$ sample from $\mathcal{N}(0, I)$
\State $\boldsymbol{x}_{\mathrm{\omega}}\gets  \sqrt{\bar{\alpha}_t}\boldsymbol{y}_0+\sqrt{1-\bar{\alpha}_t}\epsilon$ 
\For{$t = \omega$ \textbf{to} $1$}
    \State $\boldsymbol{\mu }_t, \Sigma \gets  \mu_{\theta}(x_t, t), \Sigma_{\theta}(x_t, t)$
    \State $\boldsymbol{\hat{x}}_0\gets \frac{1}{\sqrt{\alpha _t}}\boldsymbol{x}_t-\frac{\sqrt{1-\alpha _t}}{\alpha _t}\boldsymbol{\epsilon }_{\boldsymbol{\theta }}(\boldsymbol{x}_t,t)$
    \State $\boldsymbol{\hat{s}}=\frac{\left\| \boldsymbol{x}_t-\boldsymbol{x}_{t-1} \right\| _2}{\left\| \nabla _{\boldsymbol{\hat{x}}_0}\mathcal{L} \right\| _2}\cdot \boldsymbol{s}$ \hfill \textcolor{blue}{$ \rhd$ Dynamic guidance scale}
    \State $\boldsymbol{\hat{N}}\gets max\left( 1,\frac{\left\| \boldsymbol{x}_t-\boldsymbol{x}_{t-1} \right\| _2}{\left\| \nabla _{\boldsymbol{\hat{x}}_0}\mathcal{L} \right\| _2}\cdot \boldsymbol{N} \right)$\hfill \textcolor{blue}{$\rhd$ Dynamic gradient steps}
    % \If{$S_{\text{start}} \leq t \leq S_{\text{end}}$}
        \Repeat 
            \State $\boldsymbol{x}_t \gets$ sample from $\mathcal{N}(\boldsymbol{\mu }_t-\boldsymbol{\hat{s}}\Sigma \nabla _{\boldsymbol{\hat{x}}_0}\log \boldsymbol{p}_{\boldsymbol{\theta }}(\boldsymbol{y}|\boldsymbol{\hat{x}}_0),\Sigma )$
            \State $\boldsymbol{\hat{x}}_0\gets \frac{1}{\sqrt{\alpha _t}}\boldsymbol{x}_t-\frac{\sqrt{1-\alpha _t}}{\alpha _t}\boldsymbol{\epsilon }_{\boldsymbol{\theta }}(\boldsymbol{x}_t,t)$
        \Until $\boldsymbol{\hat{N}} - 1$ times
    % \EndIf
    \State $ \boldsymbol{x}_{t-1}\gets \mathcal{N}(\boldsymbol{\mu }_t-\boldsymbol{\hat{s}}\Sigma \nabla _{\boldsymbol{\hat{x}}_0}\log \boldsymbol{p}_{\theta}(\boldsymbol{y}|\boldsymbol{\hat{x}}_0),\Sigma )$
\EndFor
\State \Return $\boldsymbol{\hat{x}}_0$
\end{algorithmic}
\end{algorithm}

\textbf{Dynamic Guidance Scheme.} As illustrated in Fig.~\ref{method}, given the desired attributes, we construct the corresponding classifier and apply classifier guidance during the generative process. Similarly to the external classifier gradient guidance in~\cite{diffsuion_beat_gan}, we evaluate the anti-gradient $-\nabla_{\boldsymbol{\hat{x}}_0}\mathcal{L}$ to bring the attribute-guidance to the generative process. \textit{\textbf{However, our observations indicate that the traditional guidance scheme often leads to suboptimal outcomes.}} Concretely, the traditional guidance scheme, which adpots a constant gradient scale $\boldsymbol{s}$, often falls short in guiding the output towards the target value. Additionally, it executes merely one single gradient step per denoising step, which may not sufficiently steer the output towards the intended target, especially when early-phase denoising process intermediate outputs are significantly affected by noise. Such limitations are particularly adverse within LIE tasks that demand high similarity to the target. To mitigate this issue, we introduce a dynamic guidance scheme that consists of two distinct components, i.e., the dynamic guidance scale $\boldsymbol{\hat{s}}$ and the dynamic gradient steps $\boldsymbol{\hat{N}}$ at each denoising step. Specifically, we calculate the $\boldsymbol{\hat{s}}$ and $\boldsymbol{\hat{N}}$ based on the magnitude change of the intermediate image~\cite{voynov2023sketch,chefer2023attend}:
\begin{equation}
\boldsymbol{\hat{s}}=\frac{\left\| \boldsymbol{x}_t-\boldsymbol{x}_{t-1} \right\| _2}{\left\| \nabla _{\boldsymbol{\hat{x}}_0}\mathcal{L} \right\| _2}\cdot \boldsymbol{s}\,\:\mathrm{and} \: \boldsymbol{\hat{N}}=max\left( 1,\frac{\left\| \boldsymbol{x}_t-\boldsymbol{x}_{t-1} \right\| _2}{\left\| \nabla _{\boldsymbol{\hat{x}}_0}\mathcal{L} \right\| _2}\cdot \boldsymbol{N} \right),
\end{equation}
where $\boldsymbol{x}_{t-1}\sim \mathcal{N}\left( \boldsymbol{\mu }_{\boldsymbol{\theta }},\Sigma _{\theta} \right)$, $\boldsymbol{s}$ and $\boldsymbol{N}$ are empirically set to 1.8 and 3 in all experiments, respectively. Such a dynamic guidance scheme affords users the flexibility to adjust the strength of guidance for attributes as per their unique requirements, thereby improving overall controllability.

% which may not sufficiently steer the output toward the intended target, particularly when the intermediate outputs are laden with noise in the early phases of the denoising process.

%% file: sec/4_Experiments.tex
\begin{table*}[!t]
\centering
\caption{Quantitative comparison on LOLv1~\cite{RetinexNet}, SICE~\cite{SICE} and LOLv2-synthetic~\cite{yang2021sparse}. “T”, “S” and “U” represent “Traditional”, “Supervised” and “Unsupervised” methods, respectively. The best results of “S” and “U” are marked in \textcolor{blue}{blue} and \textcolor{orange}{orange}, respectively.}\label{table_results1}
% \resizebox{17.5cm}{!}{
\scalebox{0.8}{
\setlength\tabcolsep{2.5pt}
\renewcommand\arraystretch{1}
\begin{tabular}{c|c|ccc|ccc|ccc}
\hline 
% \rowcolor{mygray}
% \toprule
% ~ & ~ & \multicolumn{3}{c|}{LOL} & \multicolumn{3}{c|}{SICE} & \multicolumn{3}{c}{LSRW} \\ 
% \cline{3-11} 
% \rowcolor{mygray}	% \midrule \bottomrule
% \multirow{-2}*{Method} & \multirow{-2}*{Type} &
% PSNR$\uparrow$ & SSIM$\uparrow$ & LPIPS$\downarrow$ & PSNR$\uparrow$ & SSIM$\uparrow$ & LPIPS$\downarrow$ & PSNR$\uparrow$ & SSIM$\uparrow$ & LPIPS$\downarrow$  \\
% \midrule
\toprule
 \multirow{2}{*}{Method}  & \multirow{2}{*}{Type} &
 \multicolumn{3}{c|}{LOLv1}&\multicolumn{3}{c|}{SICE}&\multicolumn{3}{c}{LOLv2-synthetic}\\
 \cmidrule{3-5} \cmidrule{6-8} \cmidrule{9-11}
 % \rowcolor{mygray}
 ~ &~ &PSNR$\uparrow$ & SSIM$\uparrow$ & LPIPS$\downarrow$ & PSNR$\uparrow$ & SSIM$\uparrow$ & LPIPS$\downarrow$ & PSNR$\uparrow$ & SSIM$\uparrow$ & LPIPS$\downarrow$  \\
\midrule
SDD~\cite{SDD}& T     & 13.34 & 0.63  & 0.74  & 15.34 & 0.73  & 0.26  & 16.46 & 0.73  & 0.35 \\
LECARM~\cite{LECARM} & T     & 14.40 & 0.54  & 0.32  & 18.59 & 0.78  & 0.26  & 17.44 & 0.76  & 0.37 \\
\midrule
MBLLEN~\cite{MBLLEN}  & S     & 15.25 & 0.70  & 0.32  & 18.41 & 0.73  & 0.31  & 18.16 & 0.80  & 0.28\\
RetinexNet~\cite{RetinexNet} & S  & 17.60 & 0.64  & 0.38  & 19.57 & 0.78  & 0.27  & 17.41 & 0.67  & 0.34  \\
DSLR~\cite{DSLR} & S      & 15.20 & 0.59  & 0.32  & 14.32 & 0.68  & 0.38  & 15.80 & 0.72  & 0.25  \\
DRBN~\cite{DRBN} & S      & 19.67 & 0.82  & 0.16  & 18.73 & 0.78  & 0.28  & 21.51 & 0.82  & 0.27\\
DiffLL~\cite{jiang2023low} & S     & 26.19 & 0.85  & 0.11  & 21.33 & 0.84  & 0.22  & \textcolor{blue}{29.46} & 0.92  &  0.09  \\
PyDiff~\cite{PyramidDiff} & S      & \textcolor{blue}{27.56} &\textcolor{blue}{0.87} & \textcolor{blue}{0.10} & 21.18 & 0.83  & 0.23  & 26.13 & 0.92  & 0.08 \\
CUE~\cite{CUE} & S      & 22.67 & 0.79  & 0.20  & 20.06 & 0.82  & 0.24  & 24.47 & 0.90  & 0.12\\
Retinexformer~\cite{Retinexformer} & S & 25.15 & 0.84  & 0.13  & \textcolor{blue}{22.32}  &\textcolor{blue}{0.85}  & \textcolor{blue}{0.20}  & 25.66 & \textcolor{blue}{0.95}  & \textcolor{blue}{0.05}\\
\midrule
EnlightenGAN~\cite{Enlightengan}& U & 17.48 & 0.65  & 0.32  & 18.73 & 0.82  & 0.23  & 16.79 & 0.76  & 0.31 \\
RUAS~\cite{RUAS} & U & 16.40 & 0.49  & 0.27  & 13.21 & 0.72  & 0.43  & 16.31 & 0.65  & 0.38 \\
SCI~\cite{SCI} & U & 14.78 & 0.52  & 0.33  & 15.94 & 0.78  & 0.45  & 18.07 & 0.77  & 0.27\\
PairLIE~\cite{pairLIE}& U     & 19.46 & 0.73  & 0.24  & 21.23 & 0.83  & 0.22  & 19.12 & 0.77  & 0.23 \\
NeRCo~\cite{Neco}& U   & 19.81 & 0.73  & 0.24  & 20.73 & 0.82  & 0.23  & 19.14 & 0.74  & 0.26  \\
ZeroDCE~\cite{ZeroDCE}& U  & 14.86 & 0.55  & 0.33  & 18.67 & 0.80  & 0.26  & 17.75 & 0.83  & 0.16   \\
ZeroDCE++~\cite{Zerodcepp}& U  &15.32 & 0.56  & 0.33  & 18.65 & 0.81  & 0.27  & 17.55 & 0.83  & 0.18 \\
RRDNet~\cite{zhu2020zero}& U     & 11.38 & 0.51  & 0.36  & 13.27 & 0.68  & 0.32  & 14.85 & 0.65  & 0.24 \\
CLIP-LIT~\cite{clip-lie}& U  & 12.39 & 0.49  & 0.38  & 13.70 & 0.73  & 0.30  & 16.18 & 0.79  & 0.20 \\
GDP~\cite{GDP}& U     & 15.83 & 0.61  & 0.34  & 14.12 & 0.67  & 0.31  & 13.21 & 0.49  & 0.36 \\\midrule
AGLLDiff (Ours) & U     & \textcolor{orange}{21.81} & \textcolor{orange}{0.84} & \textcolor{orange}{0.15} & \textcolor{orange}{22.12} & \textcolor{orange}{0.84} & \textcolor{orange}{0.21} & \textcolor{orange}{21.11}& \textcolor{orange}{0.87} & \textcolor{orange}{0.13}  \\ 
\bottomrule
\end{tabular}}
\vspace{-0.4cm}
\end{table*}

% \vspace{-0.6cm}

\section{Experiments}
% In this section, we first describe implementation details, evaluation datasets, and performance criteria. Then, we present the quantitative and qualitative comparisons with state-of-theart methods. Finally, we conduct ablative experiments to validate each component.
\subsection{Implementation and Datasets}
\textbf{Inference Requirements.} The pre-trained diffusion model we employ is a $256 \times 256$ denoising network trained on the ImageNet dataset~\cite{Imagenet} provided by~\cite{diffsuion_beat_gan}. The total number of iteration steps is defaulted to 1000. We select the final 10 steps to implement the noise addition and attribute guidance. The inference process is carried out on the NVIDIA RTX 3090 GPU. 

\textbf{Testing Datasets.} We construct one synthetic dataset and seven real-world datasets for testing. The LOLv1~\cite{RetinexNet} dataset is composed of 500 low-light and normal-light image pairs and divided into 485 training pairs and 15 testing pairs. The LOLv2-synthetic~\cite{RetinexNet} dataset 
is officially divided into two parts, i.e., 900 low-light images for training and 100 low-light images for testing. The SICE benchmark collects 224 normal-light images and 783 low-light images. Each normal-light image corresponds to 2$\sim$4 low-light images. We adopt the first 50 normal-light images and the corresponding 150 low-light images for testing, and the rest (633 low-light images) for training. Moreover, we further assess our method on five commonly used real-world unpaired benchmarks: LIME~\cite{LIME}, NPE~\cite{NPE}, MEF~\cite{MEF}, DICM~\cite{DICM}, and VV~\cite{VV}. Notably, we only utilize the testing sets to evaluate our approach. Additionally, unlike some existing methods such as LLFlow~\cite{wang2022low} that adjust brightness using reference images, potentially causing biases, we follow the approaches~\cite{UHDFourICLR2023,pairLIE} and compute metrics without using any reference information to ensure fairness.
% During testing, we conduct grid search for best controlling parameters of AGLLDiff for each dataset. Detailed parameter settings are presented in the suplementary.
\begin{table*}[!t]
\centering
\caption{Quantitative comparison on DICM~\cite{DICM}, MEF~\cite{MEF}, LIME~\cite{LIME}, NPE~\cite{NPE} and VV~\cite{VV}. “T”, “S” and “U” represent “Traditional”, “Supervised”, “Unsupervised” methods, respectively. The best results of “S” and “U” are marked in \textcolor{blue}{blue} and \textcolor{orange}{orange}, respectively. BRI. denotes BRISQUE.}\label{table_results2}
% \resizebox{12cm}{!}{
\scalebox{0.73}{
\setlength\tabcolsep{1.1pt}
\renewcommand\arraystretch{1.1}
\begin{tabular}{c|clc|clc|clc|clc|clc}
\hline 
% \rowcolor{mygray}
% \toprule
% ~ & ~ & \multicolumn{3}{c|}{LOL} & \multicolumn{3}{c|}{SICE} & \multicolumn{3}{c}{LSRW} \\ 
% \cline{3-11} 
% \rowcolor{mygray}	% \midrule \bottomrule
% \multirow{-2}*{Method} & \multirow{-2}*{Type} &
% PSNR$\uparrow$ & SSIM$\uparrow$ & LPIPS$\downarrow$ & PSNR$\uparrow$ & SSIM$\uparrow$ & LPIPS$\downarrow$ & PSNR$\uparrow$ & SSIM$\uparrow$ & LPIPS$\downarrow$  \\
% \midrule
\toprule
 \multirow{2}{*}{Method}   &
 \multicolumn{3}{c|}{DICM}&\multicolumn{3}{c|}{MEF}&\multicolumn{3}{c|}{LIME}&\multicolumn{3}{c|}{NPE} &\multicolumn{3}{c}{VV}\\
 \cmidrule{2-16}
 % \rowcolor{mygray}
 ~  &NIQE$\downarrow$ & BRI.$\downarrow$ & PI   $\downarrow$ & NIQE$\downarrow$ & BRI.$\downarrow$ & PI  $\downarrow$ &  
NIQE$\downarrow$ & BRI.$\downarrow$ & PI    $\downarrow$  & 
NIQE$\downarrow$ & BRI.$\downarrow$ & PI    $\downarrow$  & 
 NIQE$\downarrow$ & BRI.$\downarrow$ & PI    $\downarrow$   \\
\midrule
SDD(T)     &4.64  & 31.74 & 4.18  & 4.52  & 38.90 & 4.12  & 4.58  & 29.75 & 3.84  & 4.64  & 37.10 & 3.72  & 3.62  & 23.46 & 3.42  \\
LECARM (T)  & 4.24  & 28.70 & 4.34  & 4.54  & 33.60 & 4.47  & 4.92  & 31.64 & 4.12  & 9.61  & 38.70 & 5.92  & 3.68  & 23.66 & 3.31  \\
\midrule
MBLLEN (S)     & 4.54  & 36.18 & 4.15  & 5.03  & 38.75 & 4.38  & 4.70  & 32.87 & 3.84  & 4.13  & 30.72 & 3.48  & 4.68  & 43.49 & 5.06  \\
RetinexNet (S)     &4.19  & 23.42 & 3.14  & 4.56  & 35.91  & 3.91  & 5.54  & 36.58 & 4.19  & 4.76  & 33.51 & 3.16  & 5.34  & 46.80 & 5.18  \\
DSLR (S)     &3.81  & 26.97 & 3.57  & 4.18  & 27.96 & 3.84  & 4.17  & 24.09 & 3.34  & 4.55  & 33.82 & 3.40  & 4.18  & 30.59 & 4.44  \\
DRBN (S)     &4.25  & 31.72 & 4.18  & 4.18  & 32.67 & 3.66  & 4.42  & 31.64 & 3.84  & 3.61  & 24.34 & 3.24  & 3.75  & 31.48 & 4.16  \\
DiffLL (S)     &\textcolor{blue}{3.70}  & 18.08 & \textcolor{blue}{3.13}  & 3.46  & 23.27 & \textcolor{blue}{2.99}  & \textcolor{blue}{3.60}  & 19.44 & 3.06  & \textcolor{blue}{3.46}  & \textcolor{blue}{14.97} & \textcolor{blue}{2.52}  & \textcolor{blue}{2.75}  & 18.53 & \textcolor{blue}{3.03}  \\
PyDiff (S)     &3.98  & 29.79 & 3.55  & 4.12  & 29.19 & 3.65  & 4.58  & 32.82 & 3.93  & 3.66  & 26.60 & 2.82  & 3.74  & 31.23 & 4.04  \\
CUE (S)     & 3.76  & 17.87 & 3.32  & 3.63  & 25.81 & 3.21  & 3.83  & 16.90 & 3.08  & 3.53  & 19.82 & 2.75  & 3.54  & 21.83 & 3.96  \\
Retinexformer (S)     & 3.72  & \textcolor{blue}{17.22} & 3.08 & \textcolor{blue}{3.44}  & \textcolor{blue}{22.01} & 3.07  & 3.86  & \textcolor{blue}{17.41} & \textcolor{blue}{2.96}  & 3.39  & 20.51 & 2.56  & 2.97  &\textcolor{blue}{16.24} & 3.35  \\
\midrule
EnlightenGAN (U)     & 3.89  & 29.23 & 3.41  & 3.56  & 25.31 & 3.25  & 4.21  & 25.34 & 3.34  & 3.66  & 24.10 & 2.95  & 3.63  & 27.79 & 3.82  \\
RUAS (U)     &6.58  & 45.06 & 5.22  & 5.40  & 41.68 & 4.57  & 5.36  & 31.62 & 4.34  & 7.12  & 46.89 & 5.61  & 4.86  & 34.03 & 4.24  \\
SCI (U)  &  4.12 & 31.64 & 3.74 & 3.63  & 24.57 & 3.23  & 4.38  & 24.85 & 3.32  & 4.12  & 27.31 & 3.41  & 5.13  & 21.45 & 3.49  \\
PairLIE (U)     & 4.13  & 28.59 & 3.69  & 4.18 & 29.54 & 3.10 & 4.51 & 25.21 & 3.26 & 4.17  & 26.22 & 3.03  & 3.66  & 25.88 & 3.69  \\
NeRCo (U)   & 3.86 & 28.34 & 3.45  &  3.53  &  23.22 &  2.97 & 3.68  & 24.49 & 2.98 &3.56 & 25.21 & 2.80 & 3.70  & 32.18 & 3.12  \\
ZeroDCE (U)     & 3.91  & 24.05 & 3.13  & 3.51  & 26.63 & 3.13 & 4.34  & 26.48 & 3.21  & 3.80  & 22.34 & 2.92  & 4.12  & 25.13 & 3.33  \\
ZeroDCE++ (U)     &3.87  & 23.40 & 3.21  & 3.49  & 23.59 & 3.12  & 4.29  & 27.24 & 3.23  & 3.81  & 22.60 & 2.95  & 3.96  & 26.11 & 3.32  \\
RRDNet (U)     & 3.81 & 23.95 & 3.21 & 3.55 & 25.92 & 3.35 & 4.35 & 35.23 & 3.81  & 3.69  & 25.51 & 3.24  & 3.65  & 29.37 & 3.28  \\
CLIP-LIT (U)     & 3.71  & 25.78 & 3.24  & 3.57  & 27.72 & 3.22  & 3.99  & 24.41 & 3.07  & 3.62  & 24.14 & 2.74  & 3.33  & 28.66 & 3.09  \\
GDP (U)     & 4.08  & 30.11 & 3.58 & 4.10 & 28.94 & 3.31 & 4.65 & 27.05 & 3.61  & 3.72  & 25.38 & 3.09  & 3.61  & 28.29 & 3.17  \\\midrule
AGLLDiff (U)     & \textcolor{orange}{3.57} & \textcolor{orange}{19.13} & \textcolor{orange}{3.07} & \textcolor{orange}{3.44} & \textcolor{orange}{23.21} & \textcolor{orange}{3.11} & \textcolor{orange}{3.64}& \textcolor{orange}{19.13} & \textcolor{orange}{2.98} & \textcolor{orange}{3.50}& \textcolor{orange}{15.13} & \textcolor{orange}{2.58} & \textcolor{orange}{3.50}& \textcolor{orange}{21.13} & \textcolor{orange}{2.77}   \\ 
\bottomrule
\end{tabular}}
\vspace{-0.6cm}
\end{table*}

\textbf{Metrics.} For the paired datasets, we adopt two distortion-based metrics: PSNR and SSIM~\cite{SSIM} to evaluate the performance of the proposed method, and also the perceptual-based metric LPIPS~\cite{LIPIS} to measure the visual quality of the enhanced results. For the other five unpaired datasets, we use three non-reference perceptual-based metrics: NIQE~\cite{NIQE}, BRISQUE~\cite{BRISQUE}, and PI~\cite{PI} for evaluation.
% \vspace{-0.05cm}

\begin{figure*}[!t]
    \centering
\setlength{\abovecaptionskip}{-0.02cm} %调整caption与图的距离
    \setlength{\belowcaptionskip}{-0.4cm}%调整caption与下文的距离
    \includegraphics[width=1\linewidth]{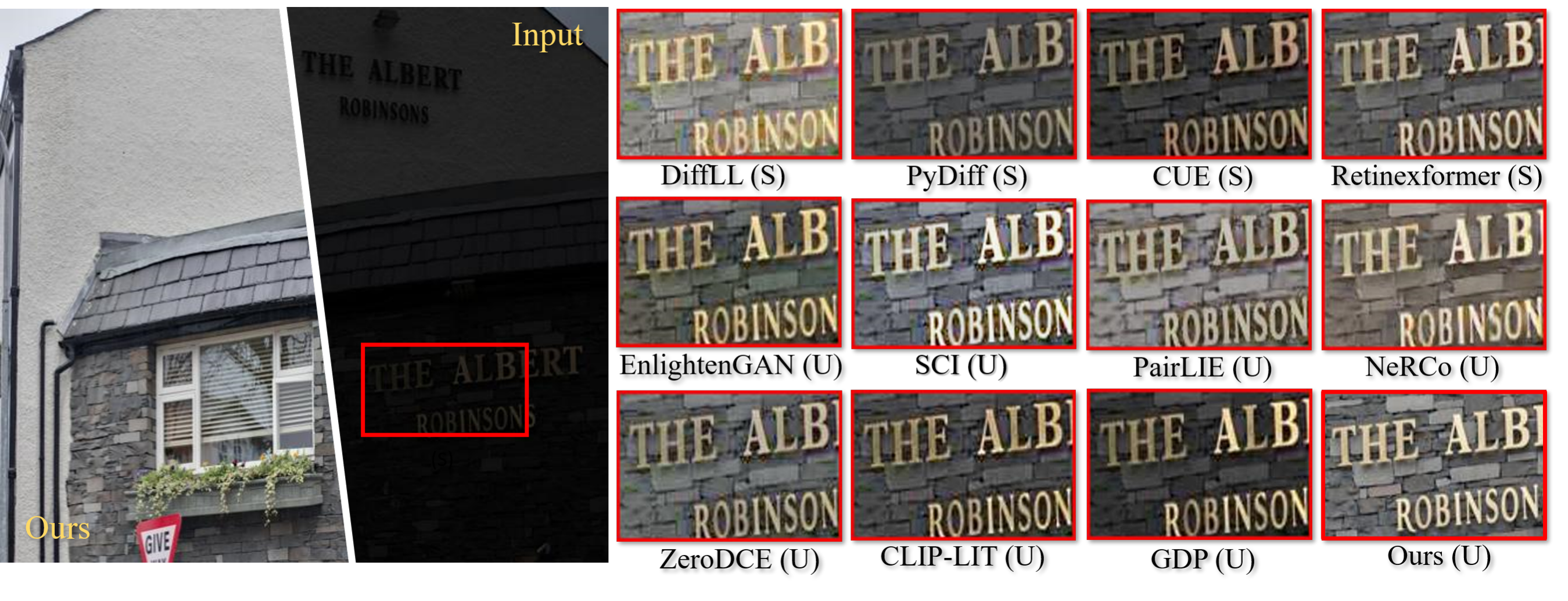}
    \caption{Visual comparisons of various LIE methods on SICE. The proposed method achieves visually pleasing results in terms of brightness, color, contrast, and naturalness.}\label{result1}
\end{figure*}

\begin{figure*}[!t]
    \centering
\setlength{\abovecaptionskip}{-0.02cm} %调整caption与图的距离
    \setlength{\belowcaptionskip}{-0.4cm}%调整caption与下文的距离
    \includegraphics[width=1\linewidth]{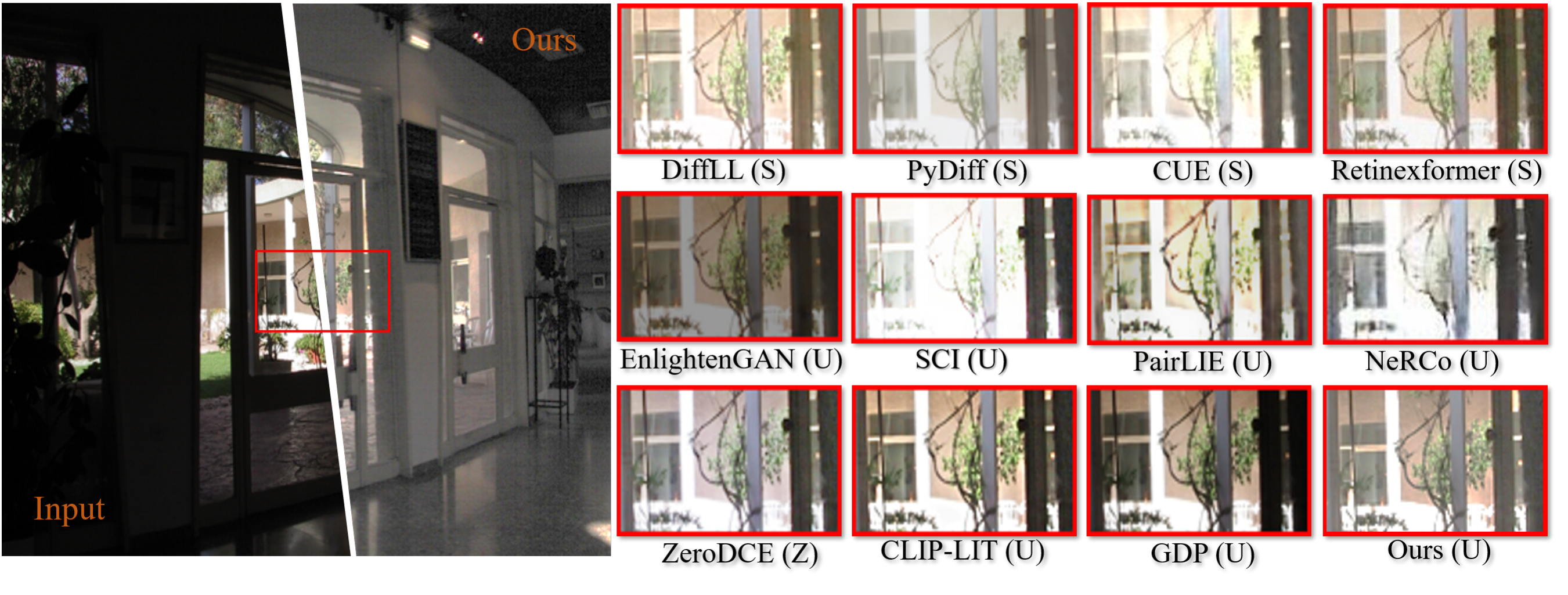}
    \caption{Visual comparisons of various LIE methods on MEF. Our method achieves remarkably higher quality among unsupervised methods with less noise and artifacts.}\label{result2}
\end{figure*}

\subsection{Comparison with the State-of-the-Art}
For a more comprehensive analysis, we compare our proposed AGLLDiff with three categories of existing state-of-the-art methods, including: 1) traditional methods SDD~\cite{SDD} and LECARM~\cite{LECARM}, 2) supervised approaches MBLLEN~\cite{MBLLEN}, RetinexNet~\cite{RetinexNet}, DSLR~\cite{DSLR}, DRBN~\cite{DRBN}, DiffLL~\cite{jiang2023low}, PyDiff~\cite{PyramidDiff}, CUE~\cite{CUE} and Retinexformer~\cite{Retinexformer}, and 3) unsupervised methods EnlightenGAN~\cite{Enlightengan}, RUAS~\cite{RUAS}, SCI~\cite{SCI}, PairLIE~\cite{pairLIE}, NeRCo~\cite{Neco}, ZeroDCE~\cite{ZeroDCE}, ZeroDCE++~\cite{Zerodcepp}, RRDNet~\cite{zhu2020zero}, CLIP-LIT~\cite{clip-lie} and GDP~\cite{GDP}. Note that the results of all those methods are reproduced by using the official codes with recommended parameters. The metrics are recalculated with the \textit{pyiqa}~\cite{pyiqa}.

% Note that the results of all those methods are reproduced by using the official codes with recommended parameters.

\textbf{Quantitative Comparisons.} Tables \ref{table_results1} and \ref{table_results2} report the quantitative performance of three paired datasets and five unpaired datasets, respectively. The best results of supervised and unsupervised methods are highlighted in blue and orange, respectively. Compared with recent competitive unsupervised approaches, AGLLDiff achieves the most advanced quantitative performance in terms of distortion-based and perceptual-based metrics across all benchmarks. Note that AGLLDiff even outperforms partial supervised approaches, which confirms the superiority of our solution.

\textbf{Visual Comparisons.}
For a more comprehensive comparison, we further provide visual comparisons with leading algorithms in Figs~\ref{result1} and \ref{result2}. Our observations are twofold: 1) the proposed method distinctly surpasses other approaches in delivering aesthetically superior enhancements in terms of brightness, color fidelity, contrast, and natural appearance, especially under extremely low-light conditions where others falter; and 2) despite supervised approaches like Retinexformer, DiffLL, and CUE exhibiting notable efficacy on LOLv1, SICE and LOLv2-synthetic datasets, their generalization capabilities may be limited as supervised learning is sensitive to the data distribution. For more visual comparisons, please refer to the supplementary material.
\vspace{-0.2cm}

\begin{figure*}[t]
    \centering
\setlength{\abovecaptionskip}{0.1cm} %调整caption与图的距离
    \setlength{\belowcaptionskip}{-0.4cm}%调整caption与下文的距离
    \includegraphics[width=1\linewidth]{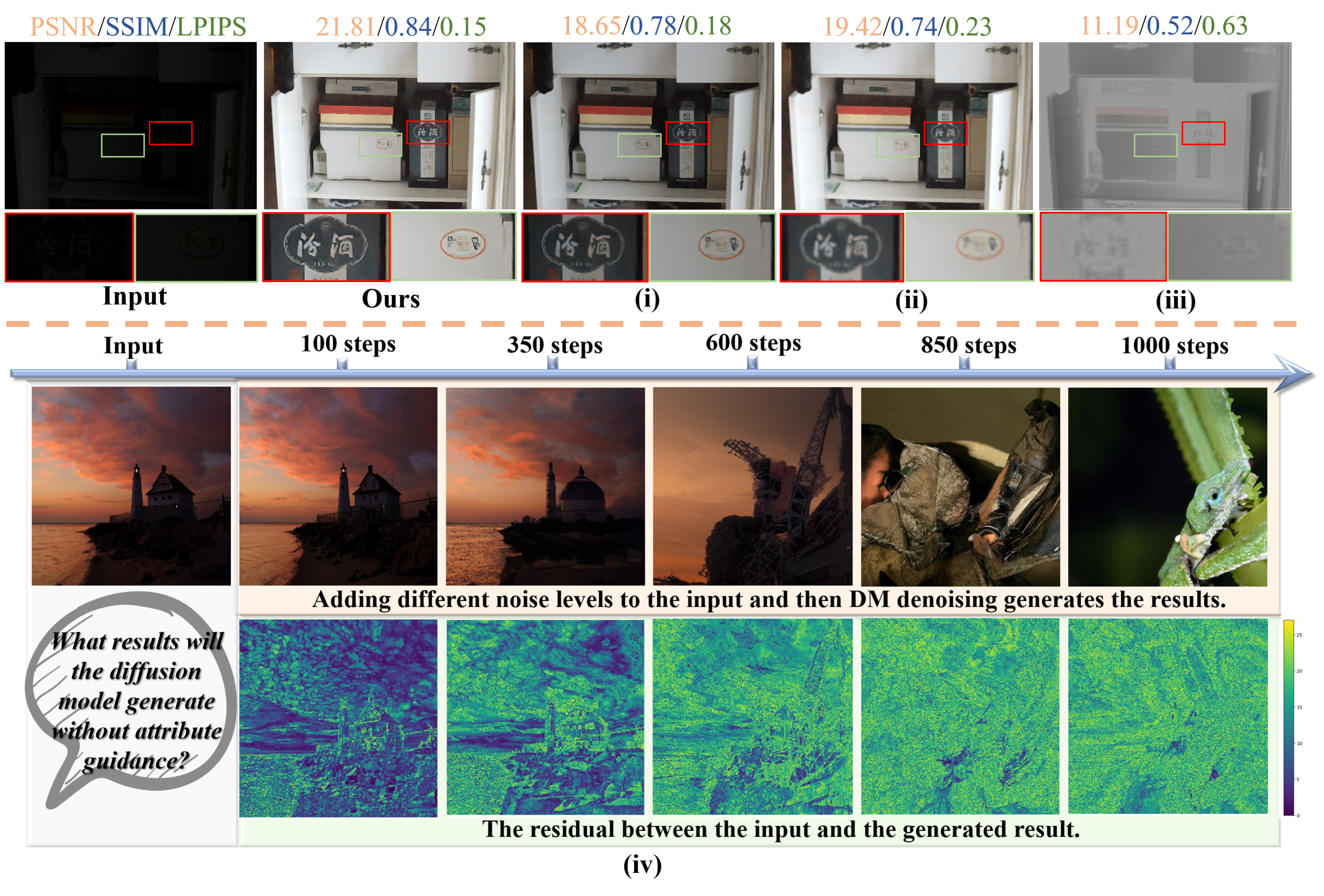}
    \caption{Visual and quantitative results of ablation studies on LOL. The full model achieves the best performance.}\label{ablations1}
\end{figure*}

\subsection{Ablation study}\label{ablation}
% To understand the role of different components of our approach, such as, the attributes, the  gradient scale $\boldsymbol{s}$, the number of gradient steps $\boldsymbol{N}$ and the iteration steps $\boldsymbol{\omega}$ of adding noise, 
To assess the impact of our approach's key components: attributes, gradient scale $\boldsymbol{s}$, number of gradient steps $\boldsymbol{N}$, and noise addition iteration steps $\boldsymbol{\omega}$, we conduct several ablation studies on the LOLv1 dataset~\cite{RetinexNet}.

\textbf{Impact of the attributes.}
We undertake ablation studies to verify the effectiveness of the mentioned attributes in Sec.~\ref{Sec3.3}. Concretely, we have tested the following three variations over the original setting: (i) without the exposure attribute guidance. (ii) without the structure attribute guidance. (iii) without the color attribute guidance. (iv) using only the pre-trained diffusion model without the attribute guidance.
Results are shown in Fig.~\ref{ablations1}. We have the following observations: 1) The removal of exposure attribute guidance limits users' control over exposure adjustments. 2) The lack of structure attribute guidance leads to blurring in the structure. 3) The absence of color attribute guidance causes severe color distortions, and the objective measures degrade significantly. 4) Without attribute guidance, relying solely on the pre-trained diffusion model, the LIE task will gradually degenerate into an unconditional image generation task as the level of noise added to the input increases.
In contrast, our full model yields clear and natural outputs, validating the efficacy of the introduced attributes.

\begin{figure*}[t]
    \centering
\setlength{\abovecaptionskip}{0.1cm} %调整caption与图的距离
    \setlength{\belowcaptionskip}{-0.4cm}%调整caption与下文的距离
    \includegraphics[width=1\linewidth]{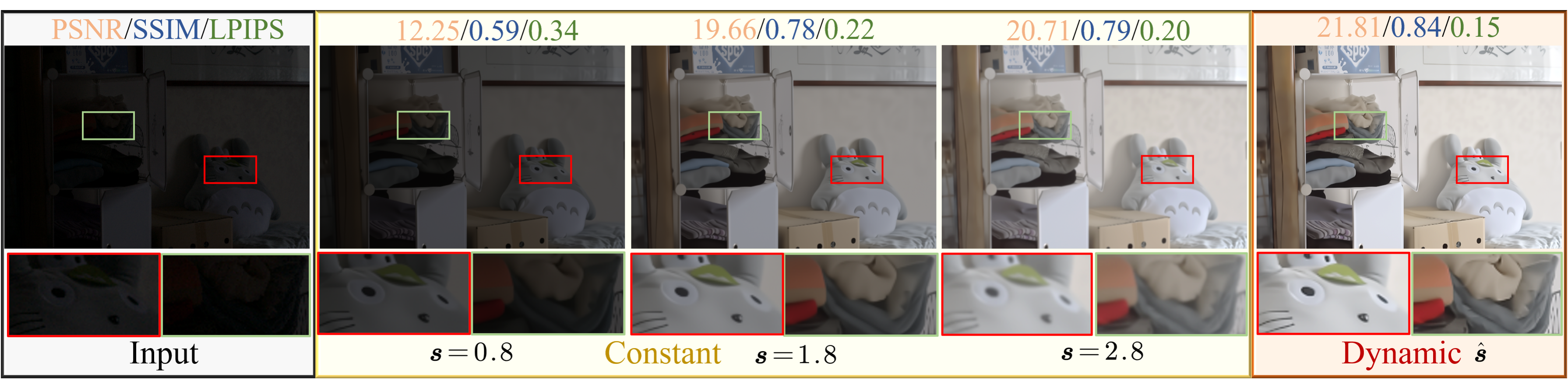}
    \caption{Ablation studies on the dynamic gradient scale $\boldsymbol{\hat{s}}$. The comparison results verify its effectiveness over the conventional constant guidance scale.}\label{ablations3}
\end{figure*}

\textbf{Effectiveness of Dynamic Guidance Scale.}
% The dynamic guidance scale $\boldsymbol{\hat{s}}$ is responsible for maintaining the fidelity of the generated output to its corresponding inputs. 
The effectiveness of the dynamic guidance scale $\boldsymbol{\hat{s}}$ is evaluated quantitatively and qualitatively. As illustrated in Fig.~\ref{ablations3}, without the dynamic guidance scale, although plausible results can be generated, the guidance scale must be manually adjusted for specific scenes, and the fidelity and clarity of the content cannot be guaranteed.
% although the generated content is plausible, the similarity and texture of the output content are obviously decreased compared to the input.
In contrast, with the dynamic guidance scale $\boldsymbol{\hat{s}}$ replacing the constant guidance scale $\boldsymbol{\hat{s}}$, high quality and fidelity results can be robustly delivered. The results highlight the indispensable role of our dynamic guidance scale in ensuring high fidelity to the target during the guidance process.
\begin{figure*}[t]
    \centering
\setlength{\abovecaptionskip}{0.1cm} %调整caption与图的距离
    \setlength{\belowcaptionskip}{-0.4cm}%调整caption与下文的距离
    \includegraphics[width=1\linewidth]{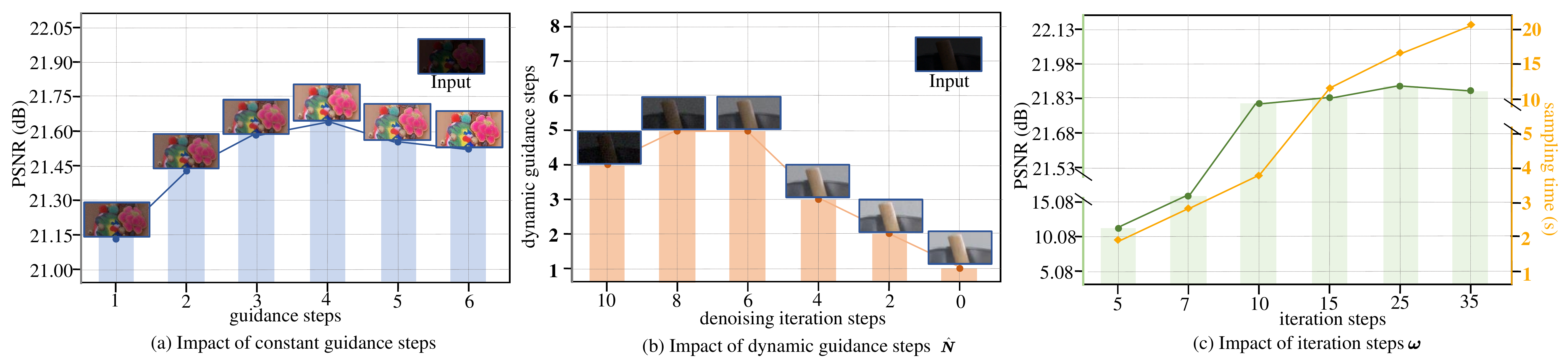}
    \caption{Ablation studies on dynamic gradient steps $\boldsymbol{\hat{N}}$ (a-b) and different iteration steps $\boldsymbol{\omega}$ (c). The blue box in (a-b) is the enhanced result.
    }\label{ablations2}
\end{figure*}

\textbf{Effectiveness of Dynamic Gradient Steps.}
The dynamic gradient steps $\boldsymbol{\hat{N}}$ serve to adaptively adjust the strength of guiding the output toward the intended target. As depicted in Fig.~\ref{ablations2}(a), employing constant gradient steps yields suboptimal results, either under- or over-enhancement, with artifacts and noise. Conversely, in Fig.~\ref{ablations2}(b), artifacts and noise are progressively removed and finer details are generated. During the early phases of the denoising process, the $\boldsymbol{\hat{N}}$ is larger, while in the later stages, $\boldsymbol{\hat{N}}$ is smaller. Such a phenomenon suggests that the intermediate outputs are laden with noise in the early phases, and hence the gradient step should be increased to effectively steer the outputs towards the intended target. Whereas in the later phases, the gradient step should be decreased to produce refined results.

% should increase the gradient steps to effectively steer the output towards the intended target.

% Such a phenomenon suggests that the intermediate outputs are laden with noise in the early phases of the denoising process.

% enlarging the gradient steps 

% image and metrics variant

\textbf{Impact of Iteration Steps.}
We explore the influence of the iteration steps $\boldsymbol{\omega}$ of adding noise and removing noise. In Fig.~\ref{ablations2}(c), we perform different numbers of iterations on the input to generate multiple noisy image outcomes. One can see that enlarging the iteration steps yields limited performance gains but significantly increases the sampling time, especially as the $\boldsymbol{\omega}$ exceeds 10. Consequently, we set the $\boldsymbol{\omega} = 10$ for the trade-off between performance and sampling time.

%% file: sec/5_Conclusions.tex
\section{Conclusion and Limitations}
% This paper introduces the Attribute Guidance Diffusion (AGLLDiff) framework to alleviate the challenges in real-world low-Light image enhancement, i.e., i) accurately modeling complex degradations is non-trivial, and ii) collecting distorted/clean image pairs is often impractical and even unavailable. AGLLDiff innovatively focuses on modeling desired high-quality image attributes such as structure, exposure, and color, which do not depend on specific assumptions about the degradation process. This attribute-based guidance facilitates the diffusion sampling process towards achieving high-quality image recovery. Extensive experimentation validates that our novel framework can realize zero-shot real-world low-light image enhancement and surpasses contemporary methods on eight challenging benchmarks. In the future, we intend to apply our solution to different restoration tasks.
This manuscript introduces an Attribute Guidance Diffusion (AGLLDiff) framework to alleviate the challenges in real-world low-light image enhancement (LIE). AGLLDiff innovatively focuses on modeling desired high-quality image attributes such as exposure, structure and color, which do not depend on specific assumptions about the degradation process. This attribute-based guidance facilitates the diffusion sampling process towards achieving high-quality image recovery. Despite outstanding quantitative and qualitative performance achieved in eight challenging LIE benchmarks, there remain areas for improvement, such as accelerating sampling via advanced techniques and exploring more underlying high-quality attributes. Furthermore, future work will extend the application of this framework to various restoration challenges.